A Detection-Gated Pipeline for Robust Glottal Area Waveform Extraction and Clinical Pathology Assessment

**Authors**: Harikrishnan Unnikrishnan[1], Rita Patel[2,3]

**Affiliations**:

[1]Orchard Robotics, San Francisco, California, USA.

[2]Department of Speech, Language, and Hearing Sciences, Indiana University, Bloomington, Indiana, USA.

[3]Department of Otolaryngology–Head and Neck Surgery, Indiana University School of Medicine, Indianapolis, Indiana, USA.

**Short Running Title:** Detection-gated glottal area segmentation

**Sources of financial support or funding:** This work was supported by the National Institutes of Health (NIH), National Institute on Deafness and Other Communication Disorders, Grant R01DC017923.

Declaration of Interest: None.

**Presentation**: Portions of this work were submitted as an abstract for presentation at the International Conference on Voice Physiology and Biomechanics (ICVPB) 2026. A non-peer reviewed preprint is available at arxiv: https://arxiv.org/abs/2603.02087

**Author to whom correspondence, reprint requests, and proofs should be directed:**

Harikrishnan Unnikrishnan, Ph.D.
Orchard Robotics, San Francisco, California, USA
Email: hari@orchard-robotics.com

## Abstract

**Objectives**: To develop and validate a fully automated two-stage modular glottal area segmentation framework for high-speed videoendoscopy (HSV), that is accurate, generalizable across imaging setups and supports assisted real-time playback. We demonstrate the clinical utility by processing the dataset to produce kinematic biomarkers that vary as a function of pathology.

**Methods:** We propose a deep-learning-based, detection-gated segmentation pipeline. A YOLOv8n glottis localizer is combined with a U-Net segmentation network; the localizer defines a tight crop so the segmenter sees a consistent field of view across datasets. The localizer also gates the output reducing spurious segmentation when the glottis is closed or hidden. Both models are trained on two expert-labeled datasets: GIRAFE ($N = 600$) and BAGLS ($N = 55{,}750$). Cross-dataset portability is evaluated by benchmarking the GIRAFE-trained models on the BAGLS test set ($N = 3{,}500$) without fine-tuning. Dice similarity coefficient (DSC) is reported and compared against published state of the art values. An exploratory clinical demonstration in 40 subjects assessed the kinematic biomarkers of glottal area to distinguish healthy from pathological function.

**Results**: In cross-dataset evaluation, the pipeline achieves DSC 0.745 (87% of the in-domain ceiling), demonstrating deployability without segmenter retraining. Moreover, the system achieves better or competitive in-distribution DSC of 0.81 (vs 0.71) and 0.856 (vs 0.887) on GIRAFE and BAGLS test sets when compared to the state of the art. In the clinical cohort, glottal area Coefficient of Variation (CV) significantly separated healthy from pathological function ($p = 0.006$, uncorrected). The pipeline runs at approximately 35 frames per second on commodity hardware supporting interactive review.

**Conclusion**: The detection-gated design yields assistive playback capable, institution-robust glottal segmentation. By reducing site-specific failure modes, it supports uniform extraction of laryngeal kinematic measures across endoscope types and acquisition settings. Code, trained weights and playback software are released at https://github.com/hari-krishnan/openglottal.

## Introduction

High-speed videoendoscopy (HSV) enables frame-by-frame observation of vocal fold vibration at several thousand frames per second, making it the gold standard for objective voice assessment in clinical laryngology.[1-4] The central derived quantity is the *Glottal Area Waveform* (GAW)—the per-frame area of the glottal opening as a function of time—from which kinematic biomarkers such as open quotient, fundamental frequency, and vibration regularity can be computed.[1, 5-7]

Early methods in glottal analysis predominantly employed active contours[8-12] and energy-based tracking[13], often requiring manual landmark initialization to navigate the complexities of laryngeal imagery. These frameworks were instrumental in establishing the first objective kinematic standards[11,14-16], with early work quantifying vibratory patterns in both clinical cohorts[15, 17] and specialized pediatric populations.[18] The methodological foundation for extending these quantitative standards to pediatric populations was significantly advanced through motion based dynamic Region of Interest and specialized processing techniques for high-speed laryngeal videoendoscopy (HSV).[18] By addressing the unique spatio-temporal constraints of younger subjects—such as smaller glottal dimensions and higher fundamental frequencies—these methods enabled the first rigorous comparisons of laryngeal kinematics between typical children and adult populations.[6]

The clinical utility of this quantitative lineage was further demonstrated using automated cycle-by-cycle analysis to differentiate pathological function. Specifically, measurements derived from the Glottal Area Waveform (GAW) and left and right lateral displacement were shown to distinguish the vibratory effects of vocal fold nodules in pediatric cohorts even when subjective visual inspection remained ambiguous.[7] Similar quantitative shifts have been observed in adult populations, where high-speed imaging has been utilized to characterize the biomechanical impacts of pathologies such as recurrent laryngeal nerve paralysis and vocal fold cysts.[19, 20]

Despite these successes, early systems remained sensitive to variable lighting, mucosal secretions, and the anatomical scale variations typical of diverse clinical environments. While morphological inpainting (InP)[21] and optical-flow variants[22] have remained competitive due to the historical scarcity of large-scale labeled data,[21] these methods generally lack spatial invariance—the ability to maintain performance regardless of the glottis's size or position within the frame—required for high-throughput, multi-institutional deployment.

The shift toward deep learning was accelerated by the release of the Benchmark for Automatic Glottis Segmentation (BAGLS).[23] This large-scale dataset enabled the development of sophisticated architectures such as the S3AR U-Net, which achieved an in-distribution Intersection over Union (IoU) of 79.97% using attention-gated modules.[24] To address temporal "flicker" common in frame-by-frame analysis, Fehling et al.[25] utilized convolutional LSTMs, while Nobel et al.[26] proposed BiGRU ensembles. Additionally the BAGLS consortium explored various re-training and knowledge distillation strategies to improve institutional generalizability.[27] Conversely, on the smaller GIRAFE dataset, high-capacity architectures like the transformer-based SwinUNetV2 (DSC: 0.62) were notably outperformed by classical InP methods (DSC: 0.71), highlighting the difficulty of training complex segmenters on limited clinical data.[21]

Most of the models typically process the entire endoscopic frame and remain susceptible to non-physiological artifacts in regions distant to the glottis.

Two gaps remain: (i) *Robustness:* Clinical recordings routinely contain frames in which the glottis is not visible (scope insertion, coughing, endoscope motion).[1] Existing segmentation models are not equipped to detect this condition and generate non-physiological artifacts in non-glottal frames, which introduces systematic errors into the resulting GAW. (ii) *Generalization:* Published methods are evaluated on a single dataset and mostly per frame based. Whether the learned representations transfer to videos from a different distribution, camera system, or patient population is largely untested.

Most existing frameworks still rely on the segmenter to perform implicit localization. We hypothesize that coupling a glottis localizer with a segmenter as a detection-gated pipeline will suppress non-physiological artifacts in non-glottal frames, improving robustness over frame-wise segmentation alone, and improve cross-dataset generalization without retraining, by constraining the segmenter to a normalized glottal region of interest.

Our contributions are: (i) A *detection gate* where localizer-based glottis detection acts as a finite-state switch—when the localizer fires, segmenter output within the detected bounding box is reported; when it does not, the previous box is held for at most 4 consecutive frames (1 ms at 4000 frames per second) and then the detection is zeroed. (ii) A *crop-zoom variant* (Localizer-Crop+Segmenter) where the detected bounding box is cropped and resized to the full segmenter canvas, providing higher effective pixel resolution at the glottal boundary and improved cross-dataset generalization.

## Methods

### Datasets

Two independent datasets are used in this study: Glottal Imaging dataset for Advanced segmentation, analysis and Facilitative playbacks Evaluation (GIRAFE)[21] and Benchmark for Automatic Glottis Segmentation (BAGLS).[23]

*GIRAFE*

The GIRAFE dataset contains 760 high-speed laryngoscopy frames ($256 \times 256$ pixels) from 65 patients (adults and children, healthy and pathological) with pixel-level glottal masks annotated by expert clinicians. Frames are grouped into official training / validation / test splits (600 / 80 / 80 frames). Splits are strictly at the patient level: the 30 training patients, 4 validation patients, and 4 test patients are disjoint sets, ensuring that no patient's anatomy appears in both training and evaluation. Each patient folder also contains the full AVI recording (median length 502 frames at 4000 frames per second) and a metadata file recording the disorder status (Healthy, Paresis, Polyps, Diplophonia, Nodules, Paralysis, Cysts, Carcinoma, Multinodular Goiter, Other, or Unknown).

*BAGLS*

BAGLS contains 55,750 training and 3,500 test frames from multiple endoscope types and institutions. These frames are typically captured at 4,000 frames per second (fps), with some instances at 2,000 fps. Image dimensions vary ($256 \times 256$ to $512 \times 512$); each frame is paired with a binary glottal mask. No patient-level labels are provided. In this study variable-size BAGLS frames are letterboxed to $256 \times 256$: i.e. the longer side is scaled to 256 pixels while maintaining aspect ratio, and the remaining dimension is zero-padded symmetrically. The same transformation is applied identically to the ground truth mask to maintain spatial correspondence. Letterboxing was chosen over non-uniform resizing to preserve the anatomical aspect ratio of the glottis, which is critical for the accurate calculation of kinematic biomarkers.

## Detection-Gated Architecture

We propose a hierarchical, two-stage framework that decouples glottal localization from pixel-level segmentation. This architecture consists of a YOLOv8n[28] object detector acting as a gate and localizer. The U-Net[29] encoder-decoder serves as the primary segmenter.

*Localizer and Temporal Consistency Guard:*

We fine-tune YOLOv8n[28] independently on each dataset. The tight upright rectangle enclosing the ground-truth mask is converted to YOLO label format for training. The default YOLOv8n augmentation pipeline is used and the BAGLS and GIRAFE are trained for 2 and 100 epochs respectively. The inputs are converted to $256 \times 256$ grayscale images. The difference in the number of epochs accounts for the smaller training split size of the GIRAFE dataset (600 frames versus 59,250). The Localizer outputs a Region Of Interest (ROI) localizing the glottal area. Moreover, the F1-score for the detector was 0.98 after 2 epochs indicating excellent learning. No ROI is expected when the glottis is closed.

For video inference, we apply a *temporal consistency model* without the memory overhead of 3D convolutions or recurrent architectures.[25] If there is no detection, the previous ROI is retained for at most 1 ms (4 frames at 4000 frames per second). Additionally, the detected box center is drift-clamped to at most 30 pixels per frame to reject spurious jumps; This temporal consistency model gates the subsequent segmenter output and removes spurious detections (e.g. stale boxes from closed glottis or scope motion) while preserving the natural opening–closing motion of the glottis.

For frame-level benchmarks such as BAGLS, where frames have no temporal order, the detector is run per frame with no temporal state. Because all temporal reasoning is confined to this gating layer, the U-Net remains a standard 2D segmenter.

*U-Net Segmenter: The Segmenter*

We train two U-Net[29] (the *segmenter*) variants with a four-level encoder–decoder (channel widths $32, 64, 128, 256$, 7.76M parameters): Full-frame U-Net, where the full frame ($256 \times 256$ grayscale) is the input and a Crop-mode U-Net where frame the detected bounding box (plus 8 px padding on each side) is cropped and resized to $256 \times 256$.

The segmenter is trained for each dataset using the respective training split, with augmentation (random flips, $\pm 30°$ rotation, $\pm 15\%$ scale jitter, brightness / contrast / Gaussian blur perturbations). Equally weighted sum of Binary Cross Entropy ($L_{BCE}$ ) and Dice Similarity Coefficient ($L_{DSC}$) was used as the loss function. The model was trained with an AdamW optimizer[30], at a learning rate of $1 \times 10^{-3}$ with cosine annealing.[31] Training runs for 20 epochs on BAGLS and 50 epochs on GIRAFE. To train Crop-mode U-Net a bounding box is derived by padding the ground truth masks and the mask undergoes the same crop-resize. The Training parameters— augmentation, loss function, and learning rate— is identical to the full-frame model.

## Processing Pipelines Evaluated

The hierarchical processing pipelines are illustrated in Figure 1. All frames are first standardized to $256 \times 256$ pixels via letterboxing and passed through the localizer. If the localizer does not detect a glottal region, the pipeline yields an all-zero mask. If a detection occurs, segmentation is performed using one of two U-Net strategies. To assess the impact of gating and resolution, we evaluate three distinct pipelines:

*Segmenter only*

Run the full-frame U-Net on the $256 \times 256$ grayscale input; output the thresholded probability map directly. No detection gate—every frame produces a prediction.

*Localizer+Segmenter*

(1) Run the detector on the full frame. (2) Run the full-frame U-Net on the full frame. (3) Zero the U-Net mask outside the detected bounding box. If the detector does not fire (or after 4 consecutive misses in the case of video), output is all-zero, removing spurious detections.

*Localizer-Crop+Segmenter*

(1) Run the detector. (2) Crop the detected region ($+8$ px padding), resize to $256 \times 256$. (3) Run the crop-mode U-Net on the resized crop. (4) Resize the output mask back to the original crop dimensions. (5) Paste into a full-frame zero mask at the detected coordinates. If the detector does not fire (or after 4 consecutive misses), output is all-zero.

## Kinematic Feature Extraction from the Glottal Area Waveform

For each patient video the Localizer + Segmenter pipeline is applied to every frame, yielding an area waveform $A(t)$. Seven scalar kinematic features, summarized in Table 1 are extracted from $A(t)$ . The fundamental frequency $f_0$ is estimated from the dominant FFT peak and converted from cycles per frame to Hz using the camera frame rate $f_s$ = 4000 frames per second. Features are compared between Healthy ($N = 15$) and Pathological ($N = 25$) groups using the two-sided Mann–Whitney $U$ test (significance threshold $\alpha = 0.05$); the 25 patients with Unknown or Other disorder status are excluded from the group comparison. We computed the coefficient of variation (CV; Area standard deviation over mean) and other kinematic features as listed in Table 1. Additionally, Fisher's exact *P* for the association between sex and diagnostic group to evaluate sex imbalance across cohorts. Because $f_0$ is sex-dependent, we additionally report Mann–Whitney comparisons stratified by sex.

## Evaluation Protocol

All three pipelines—U-Net only, Localizer + Segmenter, and Localizer-Crop + Segmenter—were first evaluated in-distribution, where each model was tested on the held-out test split of its respective training dataset. To assess cross-dataset generalization, we conducted a direct transfer test by evaluating the models trained on the smaller GIRAFE cohort on the BAGLS test set. This evaluation was performed without any fine-tuning or domain adaptation, providing a rigorous measure of how effectively the learned representations and the "Crop-Zoom" standardization transfer across the datasets captured with varying acquisition conditions, imaging hardware, and diverse subject populations.

*Controlled Component-Swap Analysis*

To isolate whether cross-dataset performance degradation was driven by localization failure or segmentation failure, we conducted a "component-swap" analysis. In this protocol, we independently exchanged the localizer (YOLOv8n) and segmenter (U-Net) modules between the GIRAFE-trained and BAGLS-trained pipelines. Specifically, we evaluated a hybrid pipeline pairing the BAGLS-trained localizer with the GIRAFE-trained segmenter.

All experiments were implemented using the PyTorch framework and executed on Apple M-series hardware using the Metal Performance Shaders (MPS) backend with a U-Net batch size of 16.

*Evaluation Metrics*

During the experiment following metrics were captured: (i) **Detection Recall (Det. Recall)**: Fraction of frames where the YOLO detector fired. Relevant for YOLO-gated pipelines; reported as 1.00 for detection-free baselines that always output a prediction. (ii) **DSC**: $2TP/(2TP + FP + FN)$, where TP, FP and FN are the number of true positives, false positives and false negatives. The DSC measures the harmonic mean of precision and sensitivity, effectively quantifying the spatial overlap between the predicted mask and the expert ground truth. It is particularly sensitive to the accuracy of the glottal boundaries. (iii) **Intersection over union** (IoU): $TP/(TP + FP + FN)$, (iv) $DSC_{0.5}$: Fraction of frames with DSC $\geq 0.5$, a clinical pass/fail threshold. This metric is a measure of robustness as a DSC of 0.5 represents the minimum overlap generally required for clinically relevant spatial assessment.

## Clinical Deployment Application

To make the pipeline usable without programming expertise, we implemented OpenGlottal, a cross-platform desktop application (PyQt5, MIT license) that wraps the detection-gated inference pipeline in an interactive viewer. Recordings are processed locally with no network transmission of patient data. Inference runs at approximately 35 frames per second on commodity hardware, permitting interactive review rather than only offline batch processing. Detector confidence and temporal-hold parameters are exposed as adjustable controls, and GIRAFE-trained and BAGLS-trained weights are selectable at load time.

Beyond the per-frame glottal area used in the in-distribution and cross-dataset benchmarks, the application computes left and right vocal-fold displacement waveforms along an automatically estimated medial axis. For each frame, the medial axis is located between the anterior commissure (AC) and posterior commissure (PC), and the left-edge and right-edge positions are sampled at a configurable point along this axis (0 = AC, 1 = PC). This produces three additional time series per recording: left displacement $L(t)$, right displacement $R(t)$, and a differential signal $L(t) - R(t)$. The sampling position can be placed at any anatomical site, allowing the displacement analysis to be localized to focal pathology — for example, positioning the probe at the mid-membranous fold to characterize nodule-associated asymmetry.[7] The kinematic features in Table 1 are computed either from the area waveform or from any selected displacement waveform.

## Results

### In-Distribution Performance (GIRAFE & BAGLS)

Table 2 compares our pipelines against published GIRAFE benchmarks. Our standalone segmenter achieved the highest accuracy ($DSC = 0.81; DSC_{0.5} = 96.2$), substantially outperforming InP ($DSC = 0.71$), SwinUNetV2 ($DSC = 0.62$). The gated Localizer+Segmenter reached a $DSC = 0.75$ with a 95% detection recall. On BAGLS (Table 3), the Localizer+Segmenter achieved the highest overall accuracy ($DSC = 0.86$; $IoU = 0.78$), reaching 96.5% of the S3AR U-Net state of the art.

### Sensitivity and Generalization

In Figure 3, we analyze the impact of temporal hold duration. Metrics exhibit a sharp increase from 0 to 4 frames before plateauing at the 1 ms threshold, suggesting the temporal gate successfully bridges the physiological closed-phase. In cross-dataset transfer (Table 4), the Localizer-Crop+Segmenter outperformed the ungated baseline ($DSC = 0.642$ vs $0.59$ at $\tau = 0.02$). The initial drop in recall observed during zero-shot transfer was partially mitigated by adjusting the detection confidence. As shown in Figure 4, a post-processing sweep of the localizer threshold reveals that the pipeline's performance on the BAGLS test set is highly sensitive to the confidence setting. Furthermore, a controlled component swap (Table 5) showed that the GIRAFE-trained crop segmenter tracks the BAGLS-trained version almost in lockstep (separation $< 0.01$ DSC). This hierarchy demonstrates that while the full-frame model exploits global context for peak in-domain accuracy, the crop-based pipeline provides superior institutional portability.

Figure 5 further details the localizer's behavior on the BAGLS dataset, specifically focusing on the "detection recall cliff". Notably, the comparison reveals that the GIRAFE-trained crop segmenter tracks the BAGLS-trained version almost in lockstep across the entire range. This demonstrates that the segmenter has learned generic laryngeal morphology rather than overfitting to center-specific imaging artifacts.

## Technical Validation: GAW Kinematics

To validate clinical utility, we extracted Glottal Area Waveform (GAW) features from the 40 subjects from the GIRAFE cohort with confirmed labels (15 Healthy, 25 Pathological) to determine if automated segmentation replicates known group differences between Healthy and Pathological voices. Example GAWs for one Healthy and two Pathological patients are illustrated in Figure 6, showing the waveform morphology extracted by the pipeline. The clinical groups were sex-imbalanced: the Healthy group was 80% female (12F/3M), while the Pathological group was 56% male (14M/11F; Fisher's exact $p = 0.025$). Consequently, results are stratified by sex (Table 6). In the female subgroup (12

Healthy vs. 11 Pathological), the coefficient of variation (CV) emerged as the primary discriminant: Healthy (Female): $0.95 \pm 0.20$; Pathological (Female): $0.57 \pm 0.29$ ; Healthy female voices exhibited significantly higher vibratory variability compared to pathological cases ($p = 0.006$). In the male subgroup (N=3 Healthy vs. N=14 Pathological), CV showed a consistent directional trend ($0.75 \pm 0.19$ vs. $0.63 \pm 0.40$) but did not reach significance ($p = 0.509$).

## Discussion

This study addresses a critical gap in automated laryngeal analysis: the inability of high-performance segmentation models to maintain accuracy when transferred across different clinical institutions and endoscopy hardware. By proposing a two-stage decoupled architecture, we demonstrate that separating localization from segmentation allows for robust performance across varying imaging conditions.

*Optimizing Segmenter baseline*

Before evaluating portability, we established a high-performance baseline. Our segmenter alone achieves a DSC of 0.81, significantly outperforming the original GIRAFE benchmark (0.64) despite using identical data splits.[21] This $+0.17$ improvement is attributable to three refinements in the training recipe: *(i) Grayscale input* (1 channel vs. 3-channel RGB)—the glottal gap is defined by intensity contrast, so color triples the input dimensionality without adding discriminative signal, making the network harder to train on only 600 frames; *(ii) Combined BCE + DSC loss* versus Dice loss only—the BCE term supplies stable per-pixel gradients that complement the region-level DSC objective and avoid the gradient instability of pure DSC near $0$ or $1$; *(iii) Higher learning* *rate with cosine annealing* ($10^{-3}$ vs. fixed $2 \times 10^{-4}$) and AdamW[30] instead of Adam, which together explore the loss landscape more aggressively and converge in 50 epochs to a stronger minimum than 200 epochs at a fixed low rate. These are engineering choices rather than architectural novelties, yet they yield a $+0.17$ DSC improvement—underscoring that on small medical-imaging datasets the training recipe matters as much as model design.

*Cross-Dataset Generalization*

The controlled component-swap analysis demonstrates that the GIRAFE-trained crop segmenter tracks the BAGLS-trained version in near-lockstep, with a performance delta of only 0.010 DSC. The hybrid model (BAGLS localizer + GIRAFE segmenter) slightly outperformed the fully in-domain crop baseline (DSC 0.745 vs. 0.735), indicating that the segmenter has learned generic laryngeal morphology rather than center-specific imaging characteristics. By reaching 87% of the in-domain ceiling without any pixel-level annotations from the target domain, the architecture provides a highly deployable baseline for institutions lacking large, annotated datasets. The slight in-distribution dip of the crop pipeline relative to the full-frame approach is attributable to marginal clipping of glottises; its advantage only emerges in out-of-distribution transfer, which is precisely the clinically relevant scenario.

*The Specialist-Generalist Trade-off and Deployment Strategy*

The full-frame pipeline is a specialist: it encodes the imaging geometry of its training set and reaches a ceiling DSC of 0.856 in-distribution but degrades when that geometry changes ($DSC = 0.55$), Table 3. The crop pipeline is a generalist: it sacrifices a small amount of in-domain accuracy in exchange for stable cross-institutional performance (DSC 0.61 vs. 0.55 for the full-frame approach out-of-distribution). For clinical deployment—where the target institution's imaging geometry is unknown—the generalist approach is the correct design choice. This trade-off motivates a practical deployment strategy: maintain a single frozen generalist segmenter and train the localizer in-domain. This hybrid approach makes laborious area ground truth data collection unnecessary. A model trained on only 600 GIRAFE frames generalizes to the heterogeneous, multi-institutional BAGLS dataset to support this claim. Furthermore, Table 5 shows that this hybrid approach achieves 87% of the in-domain ceiling, demonstrating that fine-tuning a lightweight localizer is sufficient to unlock high-fidelity segmentation on unseen institutional data without the need for new pixel-level annotations.

*Localizer Sensitivity and Domain Shift*

While the segmenter generalized robustly, the primary bottleneck in cross-dataset transfer was the localizer. The component-swap analysis quantifies this directly: replacing only the localizer with an in-domain version lifted DSC from 0.562 to 0.770—a 37% relative improvement—while the segmenter remained frozen. This confirms that cross-dataset performance degradation is primarily a localization failure, not a segmentation failure. While prior accounts[23] point to diverse laryngeal appearances as a primary challenge, our analysis suggests a nuance: the glottal anatomy itself remains a stable feature across datasets, whereas the surrounding scene geometry is the true driver of domain shift.

Detection recall was partially recovered via a single-pass confidence threshold sweep (68.8% to 85.9%), confirming that the localizer is the component most sensitive to changes in lighting and camera geometry, while the segmenter remains portable.

Detection Gating: Physiological Alignment and Clinical Safety

DSC and detection recall both plateaus at a 4-frame hold—a 1 ms window that aligns with the physiological closed-phase of the glottal cycle. This suggests the temporal gate successfully bridges transient glottal closures without over-holding, and that 1 ms is the optimal balance between suppressing transient artifacts and capturing rapid glottal dynamics.

Beyond improved segmentation metrics, the detection gate eliminates a clinical workflow burden: the GAW is zeroed when the endoscope moves off target rather than producing artifactual non-zero predictions. Without this, a clinician computing open quotient or periodicity over a full recording would need to manually identify and excise off-target frames.

Clinical Validation of GAW Features and Sex Stratification

The CV significantly distinguished healthy from pathological voices in the female subgroup ($p = 0.006$), aligning with established physiology: laryngeal pathologies increase vocal fold mass and stiffness, restricting the dynamic range of glottal oscillation. The ability of the automated pipeline to detect these signatures supports its role as a high-throughput objective tool.

Sex stratification was essential: initial group differences in $f_0$ were driven by sex composition (Fisher's exact $p = 0.025$) rather than pathology. The stratified analysis confirmed that $f_0$ does not distinguish groups within either sex, while CV survives stratification as the only feature with a significant univariate effect, in females. In the male subgroup, the small healthy sample ($N = 3$) precluded inference, though the directional trend in CV suggested discriminative potential in larger cohorts. These results should be considered exploratory pending replication on a larger, sex-balanced dataset.

Lightweight Design and Future Directions

The pipeline's computational efficiency (~35 frames per second, ~11M parameters total) makes it practical on commodity hardware without dedicated GPU infrastructure, unlike foundation model alternatives. Exploring distillation from larger models to further improve segmenter accuracy without sacrificing throughput is a natural next step, as is multi-site prospective validation with sex-balanced clinical cohorts to show CV and periodicity as robust pathology discriminants.

*Modularity and the Potential for State-of-the-art integration*

A key advantage of our two-stage decoupled architecture is its modularity; the segmenter component can be treated as a "black box" and replaced with specialized architectures as they emerge in the literature. For example, recent state of the art models like the S3AR U-Net (Montalbo, 2024) achieve an in-distribution DSC of 87.9% on BAGLS by utilizing squeeze-and-axial-attention mechanisms. While our study utilized a standard U-Net to establish the baseline, the framework is designed to be segmenter-agnostic. Integrating a model like S3AR U-Net into our pipeline as the second-stage segmenter would in principle yield major benefits. This modularity ensures that as laryngeal segmentation logic improves, our pipeline can adopt these advancements without losing the clinical safety and cross-platform reliability provided by the detection-gated architecture.

*Clinical Workflow and Directions for L/R Validation*

The deployable application (https://github.com/hari-krishnan/openglottal) addresses a practical barrier to HSV adoption: without a turnkey viewer, clinical use of high-speed segmentation has required custom scripting. Local-only processing and commodity-hardware throughput together permit use in standard clinical computing environments. The medial-probe L/R displacement feature is presented here as a demonstrated capability rather than a validated biomarker; quantitative comparison of L/R asymmetry measures against expert annotations on focal pathology cohorts — particularly the nodule and paresis subgroups where global area waveforms are known to be insensitive — is a natural next step.

Limitations

The primary technical hurdle identified in this study is the localizer's sensitivity to domain shift (e.g., changes in lighting, camera geometry, and endoscope rims). While the segmenter operates as a generalist, the localizer initially suffered a drop in recall when transferred from GIRAFE to BAGLS. However, this was somewhat recovered by doing a sweep on the detection confidence. Furthermore, because the localizer is a lightweight component (3.2M parameters), it can be adapted to a new institution using only bounding-box annotations. Crucially, labeling for localization is an order of magnitude faster and less cognitively demanding than pixel-level mask annotation. This ensures that the framework remains highly deployable, as institutional "fine-tuning" is reduced to a simple task that does not require the labor-intensive creation of new segmentation ground truth.

Regarding clinical validation, the GIRAFE cohort is relatively small ($N = 40$), with relevant labels, sex-imbalanced, and analyzed without multiple-comparison correction. The male healthy subgroup ($N = 3$) is too small for robust sex-stratified inference. Consequently, the clinical findings—such as the discriminative power of the coefficient of variation (CV)—should be viewed as exploratory. While the pipeline successfully replicated established group differences within this specific cohort, these results serve primarily as a technical "proof-of-concept" for the framework's sensitivity. They justify further investigation but cannot yet be generalized to the broader clinical population without replication in larger, sex-balanced, multi-site studies that account for a wider range of vocal pathologies and endoscopic hardware.

## Conclusion

This study shows that a two-stage decoupled architecture—separating glottal localization from pixel-level segmentation—is the key to achieving robust, cross-institutional glottis analysis. By utilizing a high-recall localizer to define a standardized region of interest (ROI), we demonstrated that a single pre-trained segmenter can maintain anatomical fidelity across different endoscopy platforms. Our cross-dataset component-swap analysis indicates that the primary barrier to institutional generalization is localization rather than segmentation: replacing only the localizer module with one trained in-domain lifted the pipeline's performance from $DSC = 0.562$ to $0.770$ (90% of the in-domain ceiling). This finding establishes a new, efficient deployment strategy for clinical centers: institutional adaptation can be achieved through lightweight bounding-box annotations rather than dense, labor-intensive segmentation masks. Technically, our standalone segmenter exceeds all published baselines on the GIRAFE benchmark ($DSC$ $0.81$, $DSC \geq 0.5 = 96.2\%$), while the full BAGLS-trained pipeline ($DSC = 0.85$) surpasses the benchmark baselines and is competitive with the state of the art. Beyond these pixel-level metrics, we provided a critical technical validation by applying the pipeline to 40 patient recordings with known labels. The automatically extracted coefficient of variation (CV) of the glottal area waveform distinguished Healthy from Pathological voices ($p = 0.006$ in the female subgroup), replicating group differences established by manual clinical experts. In clinical practice, this detection-gated framework can reduce the subjectivity inherent in manual video review. Its computational efficiency (35 frames per second) enables interactive-rate processing on commodity hardware, facilitating a data-driven workflow where kinematic biomarkers such as the Open Quotient and GAW Coefficient of Variation can be used to quantify glottal insufficiency or track longitudinal treatment outcomes. By decoupling localization from segmentation, this framework makes kinematic findings scalable across diverse endoscopic hardware and institutional boundaries.

Our work demonstrates that decoupling these tasks is essential for clinical robustness. By utilizing a dedicated localizer to define a dynamic Region of Interest (ROI), we establish a new State-of-the-art DSC of 0.81 on GIRAFE and maintain high accuracy on BAGLS without the need for incremental fine-tuning.

## Data and Code Availability

All training and evaluation scripts, trained model weights, and the GIRAFE evaluation results JSON, AI assisted playback software are available at https://github.com/hari-krishnan/openglottal. The repository README describes dataset splits (training/validation/test) for both GIRAFE and BAGLS and explains how to run the detection-gated pipeline (localizer, segmenter, and evaluation scripts). The GIRAFE dataset is freely available from https://zenodo.org/records/13773163. The BAGLS dataset (Gómez et al. 2020) is available from https://zenodo.org/records/3762320.

## Acknowledgments

We thank Andrade-Miranda et al. for making the GIRAFE dataset publicly available and Gómez et al. for the BAGLS benchmark; both datasets were essential to this work.

This work was supported by the National Institutes of Health (NIH), National Institute on Deafness and Other Communication Disorders, Grant R01DC017923.

## Declaration of Interest

The authors declare that there are no known competing financial interests or personal relationships that could have appeared to influence the work reported in this paper.

## Ethical Statement

The authors confirm that this study was conducted using only secondary, de-identified data from publicly available research benchmarks (BAGLS and GIRAFE). As the research involved the analysis of pre-existing, non-identifiable datasets and did not involve direct interaction with human subjects or the collection of private health information, it was deemed exempt from institutional review board (IRB) approval in accordance with standard ethical guidelines for secondary data analysis. The original data collection for the BAGLS and GIRAFE datasets was conducted under the ethical oversight of their respective contributing institutions, and this study adheres to their terms of use and the principles of the Declaration of Helsinki.

## Tables

**Table 1**. Kinematic features extracted from the Glottal Area Waveform (GAW) for the exploratory analysis of the consistency of the segmentation and its clinical utility. px = pixels, FFT = Fast fourier transform, Hz = Hertz.

| Feature | Description |
|---|---|
| Area Mean | Mean glottal area ($px^2$) over open frames |
| Area Standard Deviation | Standard deviation of area glottal area ($px^2$) over open frames |
| Area Range | Maximum − Minimum Area per cycle (vibratory excursion) averaged across all cycles ($px^2$) |
| Open Quotient | Fraction of each glottal cycle with area $>$ 10% of mean Area mean (unitless) |
| Fundamental Frequency | Dominant frequency from FFT (Hz) |
| Periodicity | Normalized Peak Autocorrelation at lags 0-12.5 milliseconds. (unitless) |
| Coefficient of variation | Area Standard Deviation / Area Mean (unitless) |

**Table 2.** In-distribution segmentation results on the **GIRAFE** test split (4 patients, 80 frames). Published baselines from (Andrade-Miranda, Hernández-Álvarez, and Godino-Llorente 2025). Detection Recall (Det.Recall) is the fraction of frames where the localizer detected the glottal area. Det.Recall = n/a for methods that do not include a detection stage. Dice Similarity Coefficient (DSC), Intersection over Union (IoU) are reported. (DSC ≥ 0.5) represents the fraction of frames with DSC greater than 0.5 indicating clinical utility.

| **Method** | **Det.Recall** | **DSC** | **IoU** | **DSC ≥ 0.5** |
|---|---|---|---|---|
| InP (Andrade et al 2025) | n/a | 0.71 | n/a | n/a |
| U-Net (Andrade et al 2025) | n/a | 0.64 | n/a | n/a |
| SwinUNetV2 (Andrade et al 2025) | n/a | 0.62 | n/a | n/a |
| Segmenter only (ours) | n/a | **0.81** | **0.70** | **96.2%** |
| Localizer+Segmenter (ours) | 0.95 | 0.75 | 0.64 | 88.8% |
| Localizer-Crop+Segmenter (ours) | 0.95 | 0.70 | 0.57 | 77.5% |

**Table 3.** In-distribution segmentation results on the **BAGLS** test set (3500 frames). YOLO-localizer used a detection confidence threshold (**τ** =0.35); Detection Recall (Det.Recall) is the fraction of frames where the localizer detected the glottal area. Det.Recall = n/a for methods that do not include a detection stage. Dice Similarity Coefficient (DSC), Intersection over Union (IoU) are reported. (DSC ≥ 0.5) represents the fraction of frames with DSC greater than 0.5 indicating clinical utility. Localizer+Segmenter gave the highest performance.

| Method | Det.Recall | DSC | IoU | DSC ≥ 0.5 |
|---|---|---|---|---|
| S3AR U-Net (Montalbo 2024) | n/a | 0.887 | n/a | n/a |
| Segmenter only | n/a | 0.846 | 0.77 | 94.0% |
| Localizer+Segmenter (ours) | 0.896 | **0.856** | **0.78** | **94.9%** |
| Localizer-Crop+Segmenter (ours) | 0.848 | 0.735 | 0.64 | 85.8% |

**Table 4.** Cross-dataset generalization results with localizer and segmenter trained in **GIRAFE** tested on BAGLS (N = 3,500 frames) . Final row at optimized detection confidence threshold (**τ** = 0.02). No BAGLS data used in training. Dice Similarity Coefficient (DSC), Intersection over Union (IoU) are reported. (DSC ≥ 0.5) represents the fraction of frames with DSC greater than 0.5 indicating clinical utility. Det.Recall shown as 1.00 for segmenter only (no localizer gate).

| Method | Det.Recall | DSC | IoU | DSC $\geq 0.5$ |
|---|---|---|---|---|
| Segmenter only | 1.00 | 0.59 | 0.50 | 67.1% |
| Localizer+Segmenter | 0.69 | 0.55 | 0.47 | 61.9% |
| Localizer-Crop+Segmenter | 0.69 | 0.61 | 0.53 | 70.3% |
| **Localizer-Crop+Segmenter ($\tau = 0.02$)** | **0.86** | **0.64** | **0.54** | **76.4%** |

**Table 5.** Unified performance hierarchy on BAGLS test set (3,500 frames). Comparison of gated pipelines versus full-frame segmentation using a detection confidence threshold of **τ** = 0.35. Results demonstrate that while the full-frame localizer + segmenter configuration provides the highest accuracy (Dice Similarity Coefficient [DSC] = 0.856), the proposed hybrid pipeline (BAGLS Localizer + GIRAFE segmenter) achieves competitive performance within the cropped domain (DSC = 0.745). This hybrid approach significantly outperforms the cross-domain segmenter alone (DSC = 0.588), successfully mitigating the performance gap despite the inherent Detection Recall (Det.Recall) limitations introduced by gated architectures.

| Configuration | Role | Det.Recall | DSC |
|---|---|---|---|
| BAGLS Localizer + segmenter (full frame) | In-domain ceiling | 0.896 | **0.856** |
| BAGLS segmenter only | SOTA baseline | 1.000 | 0.846 |
| **BAGLS Localizer +** BAGLS segmenter (crop) | In-domain crop | 0.848 | 0.735 |
| **BAGLS Localizer + GIRAFE** segmenter (crop) | **Proposed hybrid** | 0.856 | **0.745** |
| **GIRAFE** segmenter only | Cross-domain bound | 1.000 | 0.588 |

**Table 6.** Sex-stratified kinematic features derived from automated Glottal Area Waveforms (GAW**).** Data are presented as mean +/- standard deviation for 40 subjects from the GIRAFE cohort (15 Healthy, 25 Pathological). Features were extracted using the proposed detection-gated pipeline. Group differences were evaluated via two-sided Mann-Whitney U tests to account for the non-normal distribution of the small male healthy subgroup (n=3). The coefficient of variation significantly distinguished healthy from pathological vibratory patterns in females (p = 0.006). † Fundamental Frequency excludes recordings which had aperiodic waveforms: 1 of 11 Female-Pathological and 7 of 14 Male-Pathological subjects.

| **Features** | **Healthy** | **Pathology** | ***p* - Value** |
|---|---|---|---|
| **Female (12 Healthy/ 11 Pathology)** | | | |
| Area Mean | 125.2 ± 43.1 | 247.8 ± 204.6 | 0.230 |
| Area Standard Deviation | 112.9 ± 32.2 | 118.9 ± 96.0 | 0.406 |
| Area Range | 336.7 ± 97.6 | 375.5 ± 272.2 | 0.559 |
| Open Quotient | 0.76 ± 0.21 | 0.87 ± 0.13 | 0.192 |
| Fundamental Frequency† | 241.7 ± 34.8 | 223 ± 41.9 | 0.261 |
| Periodicity | 0.96 ± 0.01 | 0.95 ± 0.01 | 0.255 |
| Coefficient of variation | 0.95 ± 0.20 | 0.57 ± 0.29 | 0.006 |
| **Male (3 Healthy / 14 Pathology)** | | | |
| Area Mean | 192.1 ± 18.3 | 172.7 ± 94.0 | 0.768 |
| Area Standard Deviation | 142.7 ± 35.0 | 92.0 ± 66.9 | 0.197 |
| Area Range | 439.7 ± 86.7 | 343.1 ± 212.3 | 0.488 |
| Open Quotient | 0.86 ± 0.15 | 0.84 ± 0.19 | 1.000 |
| Fundamental Frequency† | 183.3 ± 75.0 | 157 ± 38.3 | 1.00 |
| Periodicity | 0.96 ± 0.00 | 0.90 ± 0.12 | 0.068 |
| Coefficient of variation | 0.75 ± 0.19 | 0.63 ± 0.40 | 0.509 |

## Figures

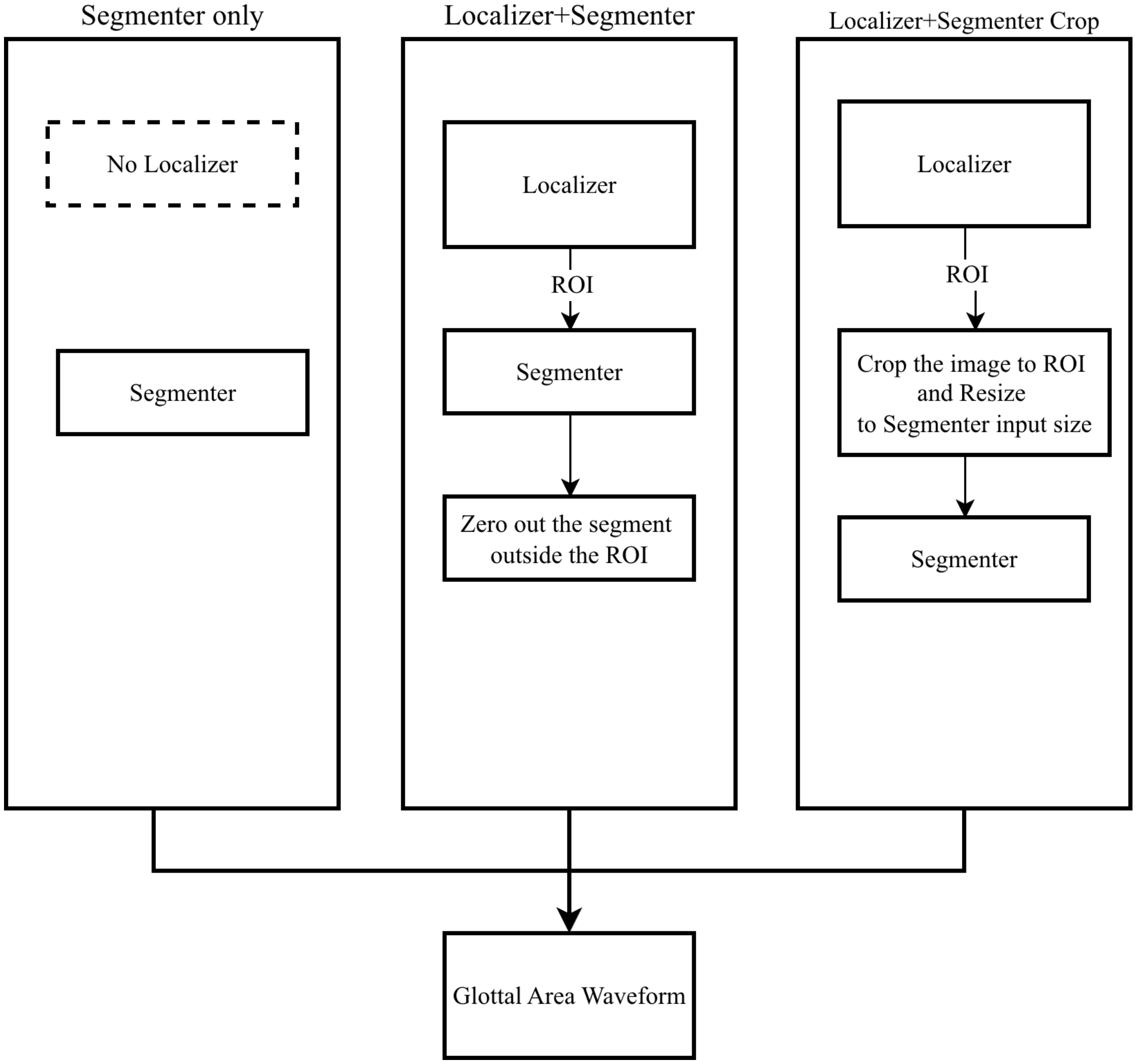

**Figure 1.** Schematic representation of the three segmentation architectures evaluated for glottal area waveform (GAW) extraction. **(Left)** *Segmenter only:* The model performing end-to-end segmentation on the full-frame image. **(Center)** *Localizer+Segmenter:* A two-stage pipeline where a localizer identifies the Region of Interest (ROI), and the segmenter output is masked to zero outside that ROI. **(Right)** *Localizer+Segmenter Crop:* A localized approach where the image is cropped and resized to the segmenter's native input resolution, maximizing spatial detail within the glottal region.

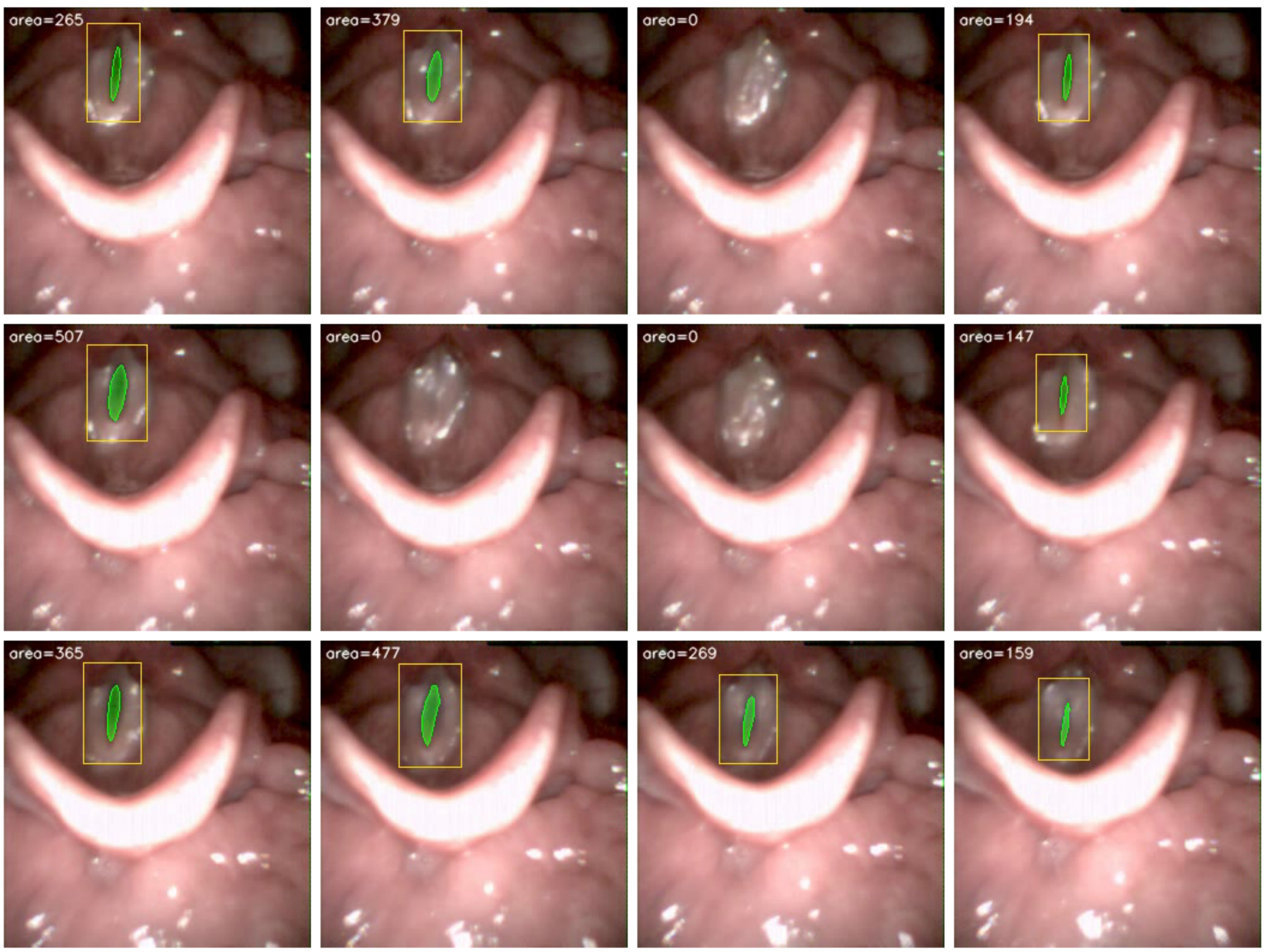

**Figure 2.** Representative output of the Localizer+Segmenter pipeline across 12 evenly spaced frames from one vibratory cycle (Participant 1). The montage displays the segmented glottal mask (green), the detection bounding box (yellow), and the calculated glottal area in square pixels (numeric labels). These frames illustrate the temporal consistency and robustness of the segmentation during the opening and closing phases.

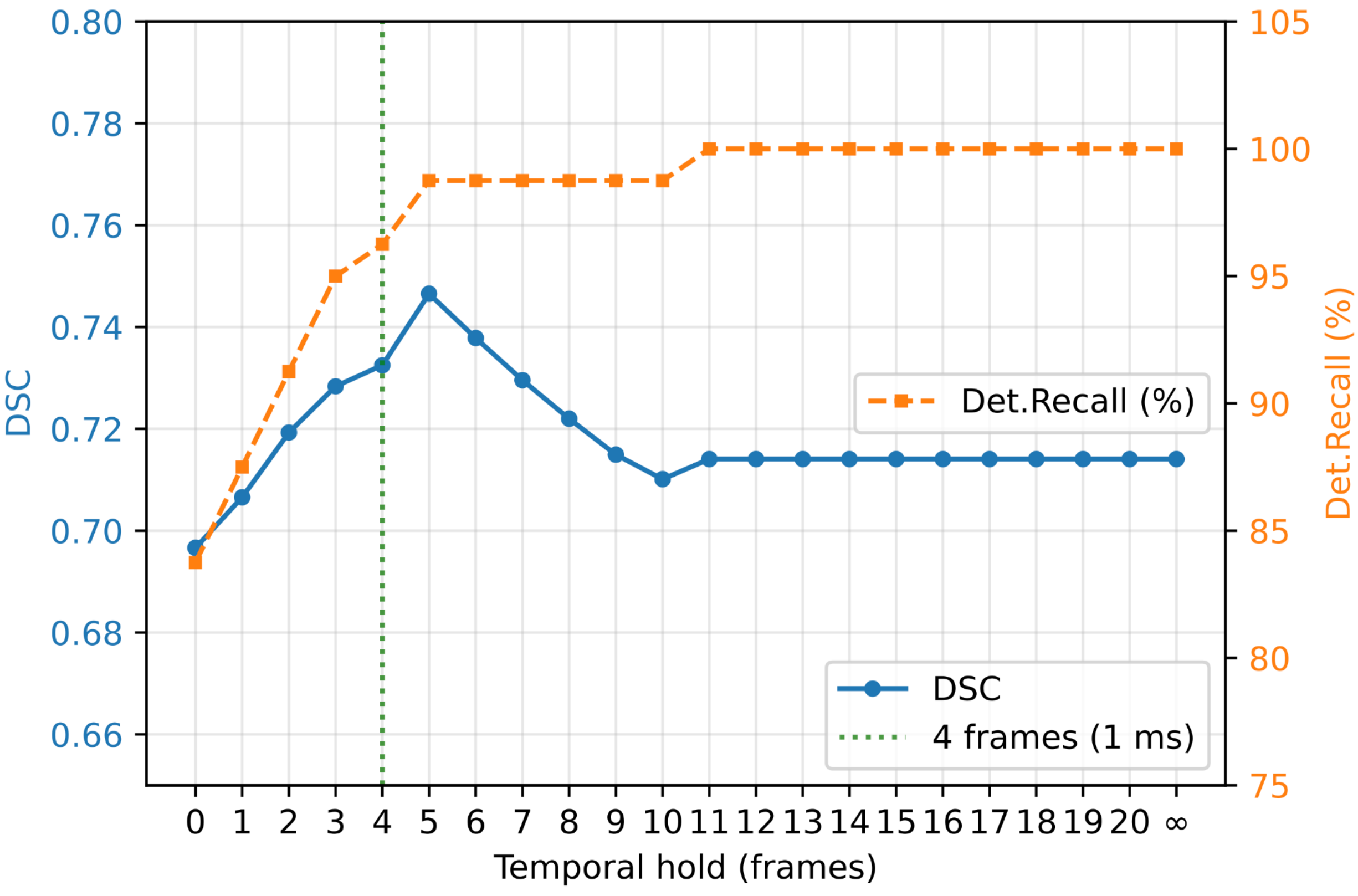

**Figure 3.** Impact of temporal hold duration on the Localizer+Segmenter (YOLO+U-Net) pipeline performance. Evaluation on the GIRAFE test set demonstrates the relationship between hold duration (0–20 frames) and dual metrics: Dice Similarity Coefficient (DSC, left y-axis) and Detection Recall (Det. Recall, right y-axis). At a 4000 fps acquisition rate, a 4-frame hold is equivalent to 1 ms. Both DSC and Det. Recall exhibit a sharp initial increase, reaching a performance plateau at approximately 4–5 frames (1 ms), suggesting that brief temporal context significantly optimizes localization and segmentation accuracy before diminishing returns occur.

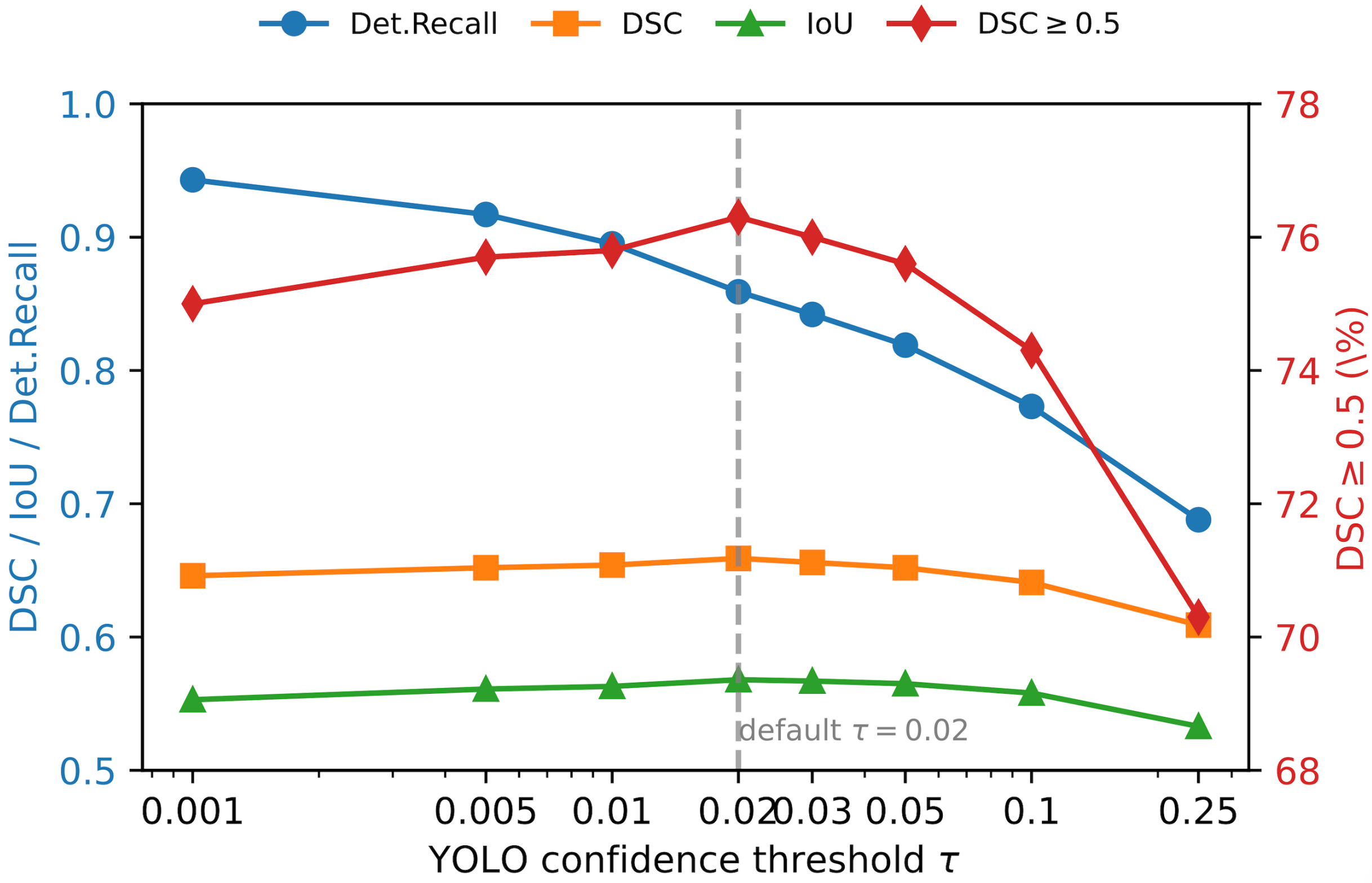

**Figure 4.** Cross-dataset sensitivity analysis of the GIRAFE-trained pipeline evaluated on the BAGLS test set. This sweep illustrates the impact of the localizer confidence threshold (**τ**) on out-of-distribution performance. While domain shift initially degrades detection recall, a significant portion of segmentation accuracy is recoverable through post-processing threshold adjustments, bypassing the requirement for additional manual annotation or site-specific re-labeling. Performance is characterized by the Dice Similarity Coefficient (DSC) and Intersection over Union (IoU), which quantify the spatial overlap between predicted and manual glottic masks. Detection Recall (Det. Recall) measures the proportion of glottic instances correctly identified by the localizer, while DSC evaluates the segmentation quality specifically within detected regions.

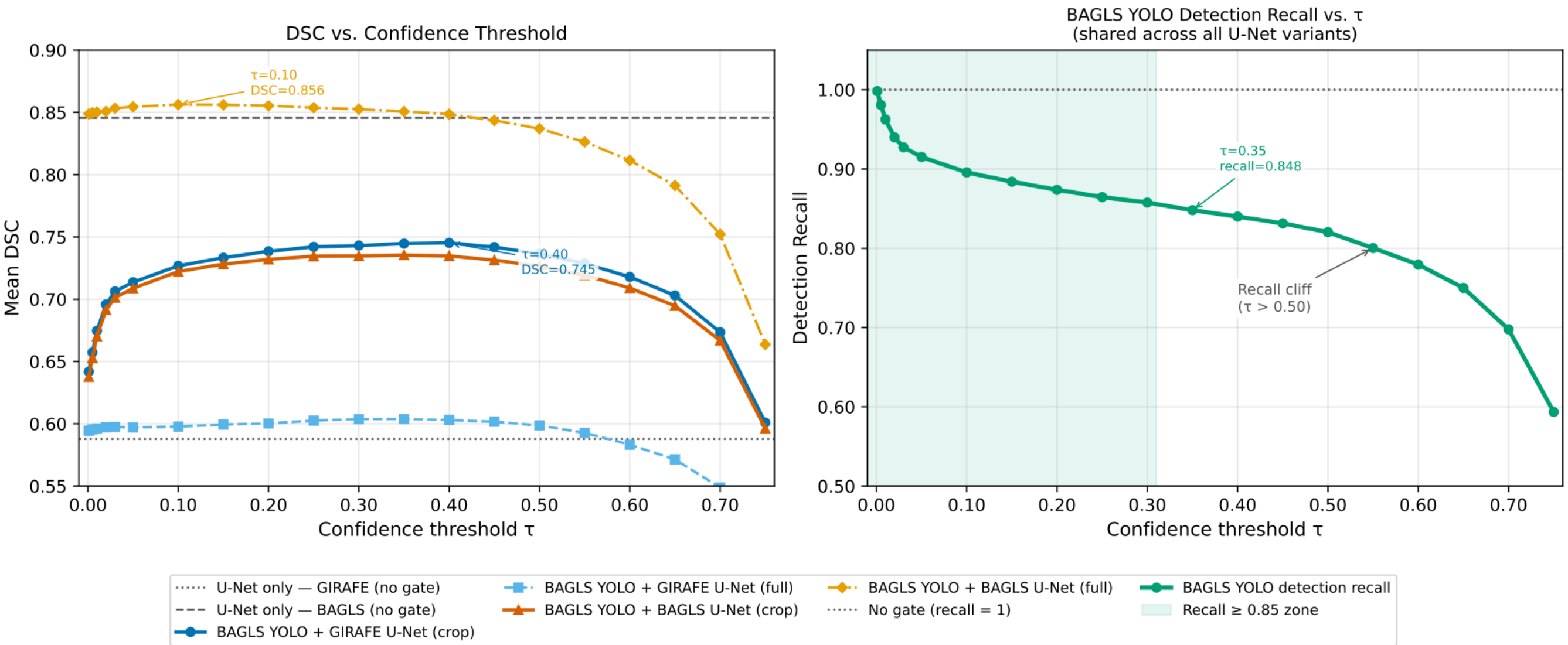

**Figure 5.** BAGLS localizer performance and segmenter consistency. Localizer confidence threshold $\tau$ sweep on BAGLS test set (3500 frames). *Left*: mean DSC for all pipeline configurations. The GIRAFE-trained crop segmenter (blue solid) tracks the BAGLS-trained crop segmenter (red solid) almost in lockstep across the full $\tau$ range, demonstrating that the segmenter has learned generic laryngeal morphology rather than centre-specific imaging artefacts. *Right*: BAGLS localizer detection recall (shared across all segmenter variants); shaded region marks. where recall ≥ 0.85, at localizer confidence threshold ($\boldsymbol{\tau}$) > 0.5 is the recall cliff and coverage degrades sharply.

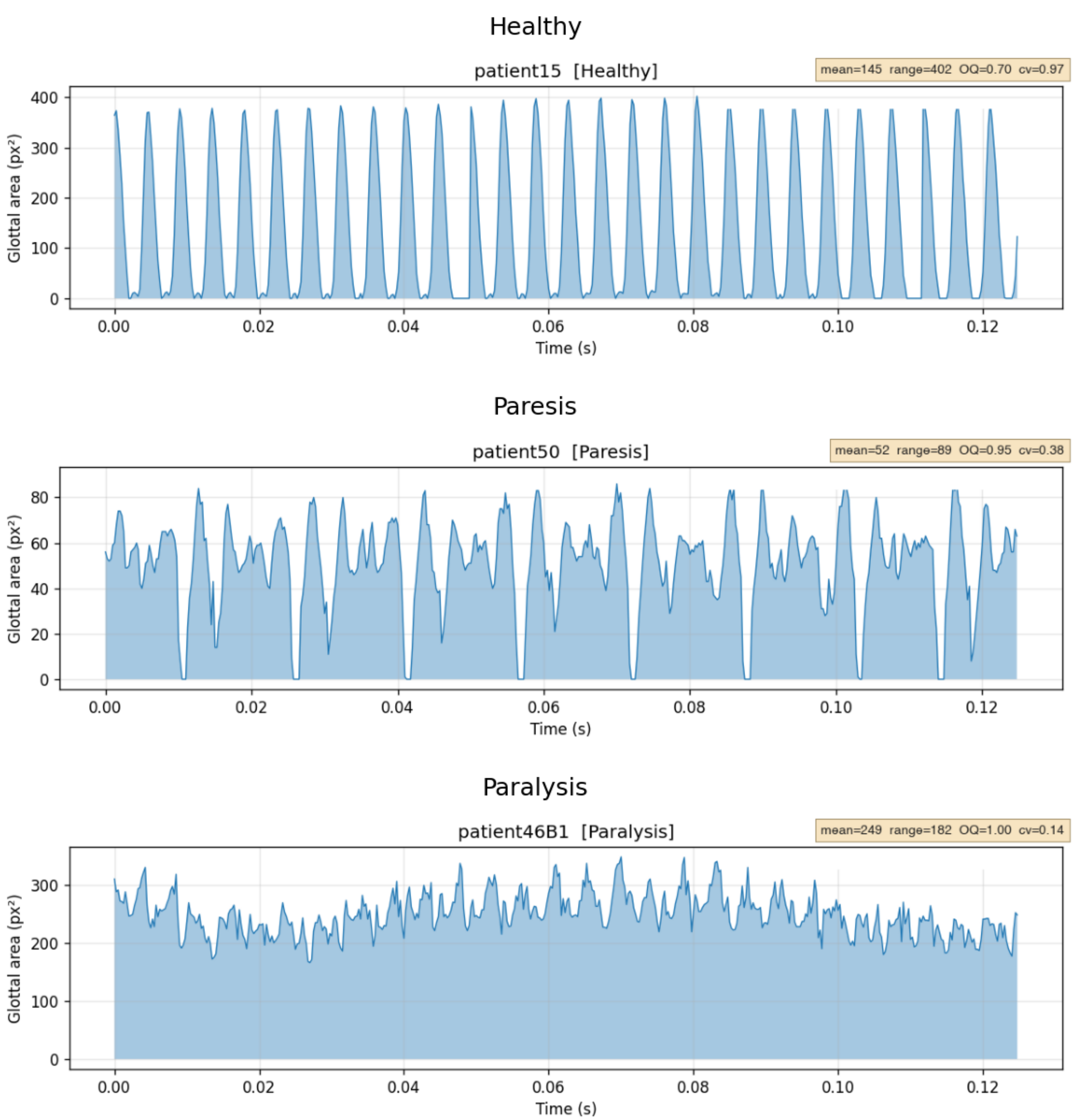

**Figure 6.** Example glottal area waveforms extracted by the 2-stage pipeline (Localizer + Segementer): Healthy (Patient 15), Paresis (Patient 50), and Paralysis (Patient 46B1). Each panel shows the time-varying glottal area extracted by the pipeline from GIRAFE raw videos.